# Encoding Data for HTM Systems

Scott Purdy
Numenta, Inc,
Redwood City, California, USA

Email: spurdy@numenta.com

We welcome comments. Please contact the author with any questions or comments.

# Encoding Data for HTM Systems

Scott Purdy
Numenta, Inc, Redwood City, California, United States of America

## Abstract

Hierarchical Temporal Memory (HTM) is a biologically inspired machine intelligence technology that mimics the architecture and processes of the neocortex. In this white paper we describe how to encode data as Sparse Distributed Representations (SDRs) for use in HTM systems. We explain several existing encoders, which are available through the open source project called NuPIC[1], and we discuss requirements for creating encoders for new types of data.

## 1. What is an Encoder?

Hierarchical Temporal Memory (HTM) provides a flexible and biologically accurate framework for solving prediction, classification, and anomaly detection problems for a broad range of data types (Hawkins and Ahmad, 2015). HTM systems require data input in the form of Sparse Distributed Representations (SDRs) (Ahmad and Hawkins, 2016). SDRs are quite different from standard computer representations, such as ASCII for text, in that meaning is encoded directly into the representation. An SDR consists of a large array of bits of which most are zeros and a few are ones. Each bit carries some semantic meaning so if two SDRs have more than a few overlapping one-bits, then those two SDRs have similar meanings.

Any data that can be converted into an SDR can be used in a wide range of applications using HTM systems. Consequently, the first step of using an HTM system is to convert a data source into an SDR using what we call an encoder. The encoder converts the native format of the data into an SDR that can be fed into an HTM system. The encoder is responsible for determining which output bits should be ones, and which should be zeros, for a given input value in such a way as to capture the important semantic characteristics of the data. Similar input values should produce highly overlapping SDRs.

## 2. The Encoding Process

The encoding process is analogous to the functions of sensory organs of humans and other animals. The cochlea, for instance, is a specialized structure that converts the frequencies and amplitudes of sounds in the environment into a sparse set of active neurons (Webster et al, 1992, Schuknecht, 1974). The basic mechanism for this process (Fig. 1) comprises a set of inner hair cells organized in a row that are sensitive to different frequencies. When an appropriate frequency of sound occurs, the hair cells stimulate neurons that send the signal into the brain. The set of neurons that are triggered in this manner comprise the encoding of the sound as a Sparse Distributed Representation.

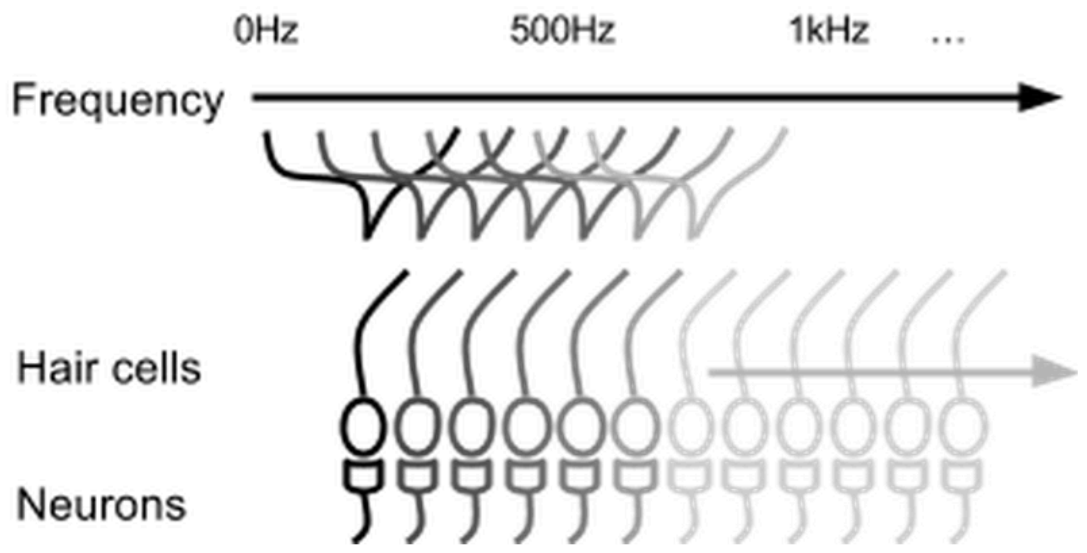

*Figure 1* Cochlear hair cells stimulate a set of neurons based on the frequency of the sound.

One important aspect of the cochlear encoding process is that each hair cell responds to a range of frequencies, and the ranges overlap with other nearby hair cells. This characteristic provides redundancy in case some hair cells are damaged but also means that a given frequency will stimulate multiple cells. And two sounds with similar frequencies will have some overlap in the cells that are stimulated. This overlap between representations is how the semantic similarity of the data is captured in the representation. It also means that the semantic meaning is distributed across a set of active cells, making the representation tolerant to noise or subsampling.

[1] http://numenta.org

The cochleae for different animals respond to different ranges of frequencies and have different resolutions for which they can distinguish differences in the frequencies. While very high frequency sounds might be important for some animals to hear precisely, they might not be useful to others. Similarly, the design of an encoder is dependent on the type of data. The encoder must capture the semantic characteristics of the data that are important for your application. Many of the encoder implementations in NuPIC take range or resolution parameters that allow them to work for a broad range of applications.

There are a few important aspects that need to be considered when encoding data:

1. Semantically similar data should result in SDRs with overlapping active bits.
2. The same input should always produce the same SDR as output.
3. The output should have the same dimensionality (total number of bits) for all inputs.
4. The output should have similar sparsity for all inputs and have enough one-bits to handle noise and subsampling.

In following sections we will examine each of these characteristics in detail and then describe how you can encode several different types of data. Note that several SDR encoders exist already and most people will not need to create their own. Those who do should carefully consider the above criteria.

1) Semantically similar data should result in SDRs with overlapping active bits

To create an effective encoder, you must understand the aspects of your data that should contribute to similarity. In the cochlea example above, the encoder was designed to make sounds with similar pitch have similar representations but did not take into account how loud the sounds were, which would require a different approach.

The first step to designing an encoder is to determine each of the aspects of the data that you want to capture. For sound, the key features may be pitch and amplitude; for dates, it may be whether or not it is a weekend.

The encoder should create representations that overlap for inputs that are similar in one or more of the characteristics of the data that were chosen. So for weekend encoders, dates that fall on Saturdays and Sundays should overlap with each other, but not as much or at all with dates that fall on weekdays.

## Preserving Semantics: A Formal Description

Here we formalize the encoding process by defining a set of rules that relate the semantic similarity of two inputs with the number of overlapping one-bits in the corresponding encoded SDRs.

Let $\mathcal{A}$ be an arbitrary input space and let $S(n,k)$ be the set of SDRs of length $n$ with $k$ ON bits. An encoder $f$ is simply a function $f : \mathcal{A} \to S(n,k)$. A distance score $d_{\mathcal{A}}$ over space $\mathcal{A}$ is a function $d_{\mathcal{A}} : \mathcal{A} \times \mathcal{A} \to \mathbb{R}$ that satisfies three conditions:

1. $\forall x, y \in \mathcal{A}, d_{\mathcal{A}}(x,y) \geq 0$
2. $\forall x, y \in \mathcal{A}, d_{\mathcal{A}}(x,y) = d_{\mathcal{A}}(y,x)$
3. $\forall x \in \mathcal{A}, d_{\mathcal{A}}(x,x) = 0$

Equation 1 requires the semantic similarity metric give a distance value of zero or greater. Equation 2 requires the distance metric to be symmetric. And Equation 3 requires that the distance between two identical values be zero.

Given an input space and a distance score, we can evaluate an encoder by comparing the distance scores of pairs of inputs with the overlaps of their encodings. Two inputs that have low distance scores should have SDRs with high overlap, and vice versa. Moreover, if two SDRs have higher overlap than two other SDRs, then the former's pre-encoding distance score should be lower than the latter's. We state this formally below.

For SDRs $s$ and $t$ with the same length, let $O(s,t)$ be the number of overlapping bits (i.e. the number of ON bits in $s\&t$). Then for an encoder $f : \mathcal{A} \to S(n,k)$ and $\forall w, x, y, z \in \mathcal{A}$,

4. $O(f(w), f(x)) \geq O(f(y), f(z)) \Leftrightarrow d_{\mathcal{A}}(w,x) \leq d_{\mathcal{A}}(y,z)$

Equation 4 states that encodings with more overlapping one bits means the values have greater semantic similarity and, inversely, that values with greater semantic similarity will have encodings with more overlapping one bits. It is not always possible to create an encoder that satisfies this, but the equation can be used as a heuristic to evaluate the quality of an encoder.

2) The same input should always produce the same SDR as output

Encoders should be deterministic so that the same input produces the same output every time. Without this property, the sequences learned in an HTM system will become obsolete as the encoded representations for values change. Avoid creating encoders with random or adaptive elements.

It can be tempting to create adaptive encoders that adjust representations to handle input data with an unknown range. There is a way to design an encoder to handle this case without changing the representations of inputs that is described below in the section labeled “A more flexible encoder method”. This method allows encoders to handle input with unbounded or unknown ranges.

3) The output should have the same dimensionality (total number of bits) for all inputs

The output of an encoder must always produce the same number of bits for each of its inputs. SDRs are compared and operated on using a bit-by-bit assumption such that a bit with a certain “meaning” is always in the same position. If the encoders produced varying bit lengths for the SDRs, comparisons and other operations would not be possible.

4) The output should have similar sparsity for all inputs and have enough one-bits to handle noise and subsampling

The sparsity for encoders can vary from around 1% to 35% but should be relatively fixed for a given application of an encoder. While keeping the sparsity the same should be the rule, small variations in sparsity will not have a negative effect.

Additionally, there must be enough one-bits to handle noise and subsampling. A general rule of thumb is to have at least 20-25 one bits. Encoders that produce representations with fewer than 20 one bits do not work well in HTM systems since they may become extremely susceptible to errors due to small amounts of noise or non-determinism.

## 3. Example 1 – Encoding Numbers

One of the most common data types to encode is numbers. This could be a numeric value of any kind – 82 degrees, $145 in sales, 34% of capacity, etc. The sections below describe increasingly-advanced encoders for a single numeric value. In each section, we will change the semantic characteristics that we desire to capture and update our encoding to achieve the new goal.

### 3.1 A Simple Encoder for Numbers

In the simplest approach, we can mimic how the cochlea encodes frequency. The cochlea has hair cells that respond to different but overlapping frequency ranges. A specific frequency will stimulate some set of these cells. We can mimic this process by encoding overlapping ranges of real values as active bits (Fig. 2). So the first bit may be active for the values 0 to 5, the next for 0.5 to 5.5, and so on. If we choose for our encoding to have 100 total bits, then the last bit will represent 49.5 to 54.5. The entire range of values that are represented will be 0 to 54.5. The minimum and maximum values in this approach are fixed. Values above or below the allowed range will be given a representation matching the maximum or minimum possible representations.

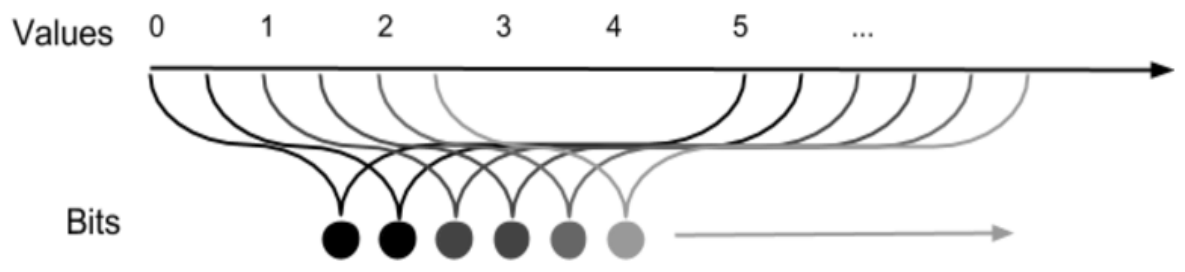

*Figure 2* Each bit in the representation responds to a range of values that overlaps with its neighbors.

Using the encoder parameters from before, here is what the encoding of the values 7.0 (Fig. 3A), 10.0 (Fig. 3B), and 13.0 (Fig. 3C) would look like. Note that numbers that are close together (7.0 and 10.0, or 10.0 and 13.0) share one bits with each other, but numbers that are not so close together (7.0 and 13.0) do not share any one bits.

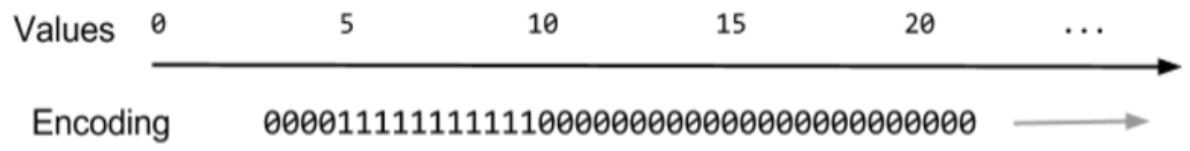

*Figure 1A* The encoded representation of 7.0 using encoding parameters of 100 total bits, minimum-value 0, value range per bit of 5.0, and increase per bit of 0.5.

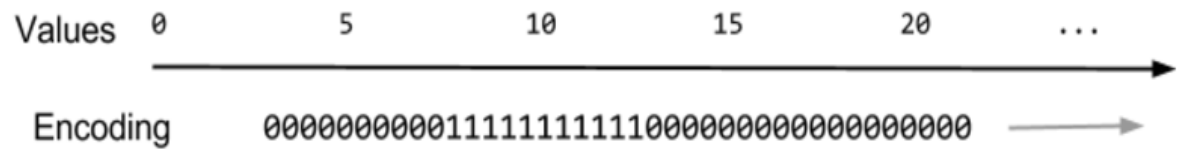

*Figure 3B* The encoded representation of 10.0 using encoding parameters of 100 total bits, minimum-value 0, value range per bit of 5.0, and increase per bit of 0.5. There are several shared one bits with the representation for 7.0.

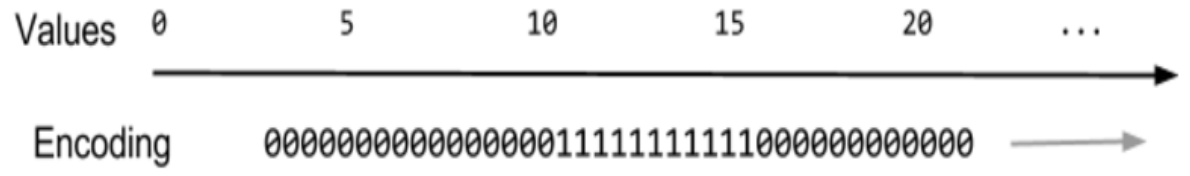

*Figure 3C* The encoded representation of 13.0 using encoding parameters of 100 total bits, minimum-value 0, value range per bit of 5.0, and increase per bit of 0.5. There are several shared one bits with the representation for 10.0, but none with the representation for 7.0.

When we create an implementation of this encoder, we first split the range of values into buckets, and then map the buckets to a set of active cells. Here are the steps for encoding a value with this approach:

1. Choose the range of values that you want to be able to represent, minVal and maxVal.
2. Compute the range as $range = maxVal - minVal$
3. Choose a number of buckets into which you will split the values.
4. Choose the number of active bits to have in each representation, w.
5. Compute the total number of bits, n: $n = buckets + w - 1$
6. For a given value, v, determine the bucket, i, that it falls into: $i = floor[buckets * (v - minVal)/range]$
7. Create the encoded representation by starting with n unset bits and then set the w consecutive bits starting at index i to active.

Note that this encoding scheme has four parameters: minimum value, maximum value, number of buckets, and number of active bits (w). Alternatively, you may choose the total number of bits, n, rather than the number of buckets, which would then be computed as $buckets = n - w + 1$.

Here is an example of encoding the outside temperature for a location where the temperature varies between 0°F and 100°F.

1. This minVal is 0°F and the maxVal is 100°F.
2. The range is 100.
3. We choose to split the range into 100 buckets.
4. We choose to have 21 active bits for each representation.
5. The total number of bits is computed to be $n = buckets + w - 1 = 120$
6. Now we can select the bucket for the value 72°F as $i = floor[100 * (72 - 0)/100] = 72$
7. And the representation will be 120 bits with 21 consecutive active bits starting at the 72nd bit:

000000000 … 000011111111111111111111100000000 … 00000000
72nd bit ↗

This encoding approach is simple and provides quite a bit of flexibility but requires that you know the appropriate range of the data. If your data falls outside the minimum and maximum values then the encoder doesn't work well. Typically you would use the smallest bucket for values below the range and the largest bucket for values above the range. Thus all values above the range will have the same representation, and similarly for those below the range. Below is the representation for a temperature of 100°F. This will also be the representation for 110°F or any other number larger than the maximum value of 100°F.

000000000000000 … 000000000000000000011111111111111111

100th bit ↗

The number of buckets can be chosen depending on the inherent noise for this metric in your application and the quality of the desired predictions. For a very noisy signal you might want a smaller number of buckets. This would allow the HTM to see more stable inputs but it would not be able to make very precise predictions. For a very clean signal you could have a large number of buckets. In this case the HTM would be able to make very precise predictions.

## 3.2 A More Flexible Encoder for Numbers

Biological sensory organs such as the cochlea have a fixed range of values they can encode. In humans this is approximately 20 hertz to 20 kilohertz. If sounds in the world shifted to higher frequencies then our ears would be useless. The encoder mechanism we described above also has a fixed range. The designer of the encoder selects the range, but once it is chosen and the system starts learning then you can't change it. However, there is a way to overcome this limit. It is possible to design encoders that have a fixed number of bits but can encode an essentially limitless range of values by using a hash function[2]. With this design, each bit in an SDR can represent multiple ranges of values. If these ranges are assigned via a hash function then the SDRs for two values that are far apart may overlap by one or two bits but this small overlap will not cause a problem for the HTM. We will now show how this works in a numeric encoder, but the same principle is used in the geospatial encoder described later in the chapter.

If we look at step 6 in the previous section, we see that each bucket is identified by a specific bit. We then select bits for the following w-1 buckets to complete the representation. Each bucket has overlapping bits with its neighbors. For a more flexible numeric encoder, we can do the same steps 1-6 but change how we select the bits in the representation. Specifically, we can use a hashing function to deterministically select one of the output bits from the bucket index. We do this selection separately for each bit, so rather than w consecutive bits being active, there are w bits active based on the hashing function. The hashing operation looks like this:

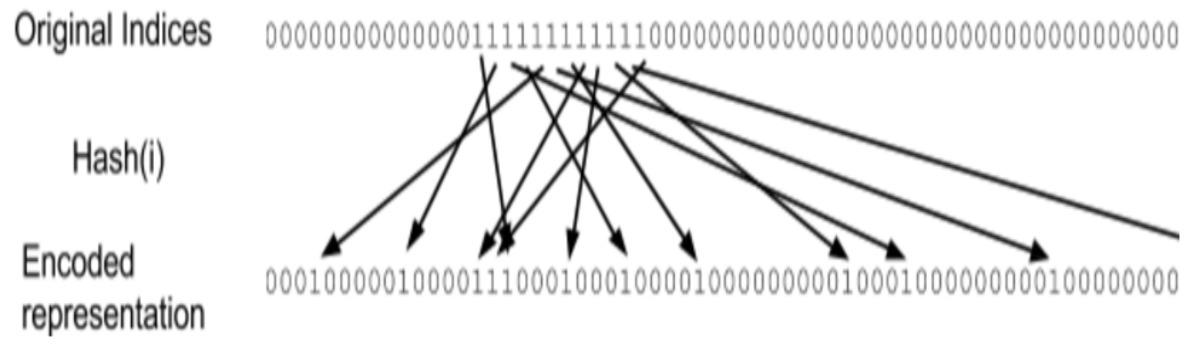

The advantage of this method is that you don't need to restrict the values to an overall range. Instead, you just need to select the range for which each bucket is responsible. It is possible that there will be some unwanted collisions. In other words, two buckets that should not have overlapping bits may actually have one or two overlapping bits due to chance. This issue is not a problem in practice as long as you choose your hash function carefully and have large enough w and n values.

If you take a single bit from this representation, you cannot know what the original value is since there are many different values that will produce encodings that include that bit. But if you take all of the active bits together, you have a representation very specific to your input. This may differ from many biological encoders, but it will work well because it satisfies the encoding properties that we outlined previously.

## 3.3 The Numeric Log Encoder

Some applications may benefit from numeric encoders that capture similarity between numbers differently based on how large the number is. In other words, the values 4000 and 5000 may be treated just as similar as the numbers 4 and 5. An encoding approach using a logarithmic function would look something like this:

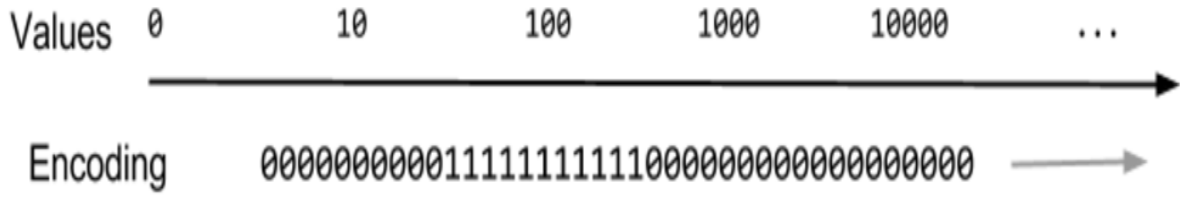

This encoder is sensitive to small changes for small numbers. A change from 3 to 15 will result in a fairly different encoding. But the encoding for larger numbers is much less sensitive. A change from 1000 to 2000, for instance, would have a much smaller impact on the encoding despite being a much larger absolute change in the number.

## 3.3 The Delta Encoder

A delta encoder is designed to capture the semantics of the change in a value rather than the value itself. This technique is useful for modeling data that has patterns that can occur in different value ranges and may be helpful to use in conjunction with a regular numeric encoder. A simple example of where you might use a delta encoder is when trying to predict a value that is constantly increasing. The predictable pattern in data with this characteristic is not the values themselves, but the change in the values – is the value larger or smaller than the previous value, and by how much? While a regular numeric encoder would give different representations for each new value, the delta encoder would produce overlapping encodings for values that increased or decreased from the previous value by a similar amount.

If you had a temperature sensor on a piece of machinery, the delta encoder would recognize temperature patterns caused by machine behavior even if the absolute range was different due to the surrounding weather patterns. The outside temperature may be 60 degrees one day and 70 the next. A regular numeric encoder's representations for values in the 60s would not share much, if any, overlap with the representations for values in the 70s. But the delta encoder would produce similar encodings when the values change in a similar way, even if the values themselves have never been seen before.

[2] https://en.wikipedia.org/wiki/Hash_function

This encoder is different from the previous encoders since it uses both the current and the previous inputs to determine the output. The implementation of this encoder is the same as one of the other scalar encoders but you apply the encoder to the difference between the current input value and the previous.

## 4. Example 2 – Encoding Categories

Many datasets contain categorical information. In some cases the data consists of discrete, completely unrelated categories (such as the SKUs of products in a store). In other cases, the data consists of categories that may have some relation (such as days of a week). The examples below show how to encode these types of data.

Some characteristics of dates and times are categorical in nature. For instance:

- Weekday vs weekend
- Holiday vs non-holiday
- Day vs night
- Meal time vs not meal time
- Part of speech for a word

In many cases it is useful to encode these characteristics as completely discrete categories. The encoding in these cases should attempt to minimize overlap between any of the category encodings. The easiest way to do this is dedicate some number of bits to each option. The encoding for any option has its dedicated bits active and the rest inactive. Here is an example of the weekday/weekend encoding for Saturday:

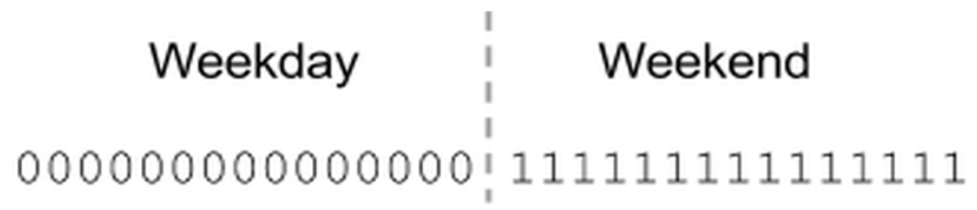

This encoding is useful to add to data streams in which the patterns in the stream are different during the week than on weekends. Adding this encoding ensures that the HTM systems receive distinct input patterns on weekdays vs. weekends, allowing them to more easily learn separate predictions for the two periods.

The category encoding can also be applied to part of speech. Here is what one such encoding may look like for a verb:

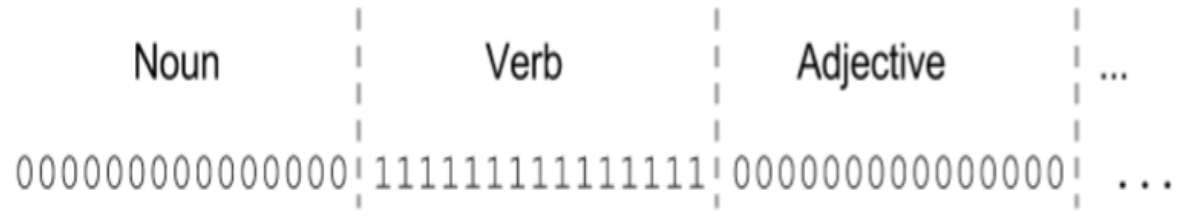

### 4.1 Ordered and Cyclic Categories

Sometimes the patterns change continuously rather than having completely discrete categories.

The difference, then, may not have a hard, discrete line. The patterns at noon may sometimes resemble the morning patterns but other days may be more similar to the common evening patterns. In these cases, you simply convert to the numeric value and use a scalar encoder to generate the SDR.

However, particularly in the case of dates and times, the categories may be cyclic in addition to continuous. For example, Friday is a weekday, but a bit different than Thursday. Sunday evening is a weekend, but not quite like Saturday evening. You can start by representing the day of the week as an integer from 0 to 6, where 0 is Sunday and 6 is Saturday. A numeric encoder would create a good representation, except that there would be little or no overlap in the encodings for Saturday and Sunday since they are on opposite ends of the range. In these cyclic cases, the encoding must "wrap." For this example, we will use a small number of bits in the encoding. In a real implementation you would want more bits active and would need more total bits as a result. The easiest way to understand this is with diagrams (Fig. 4):

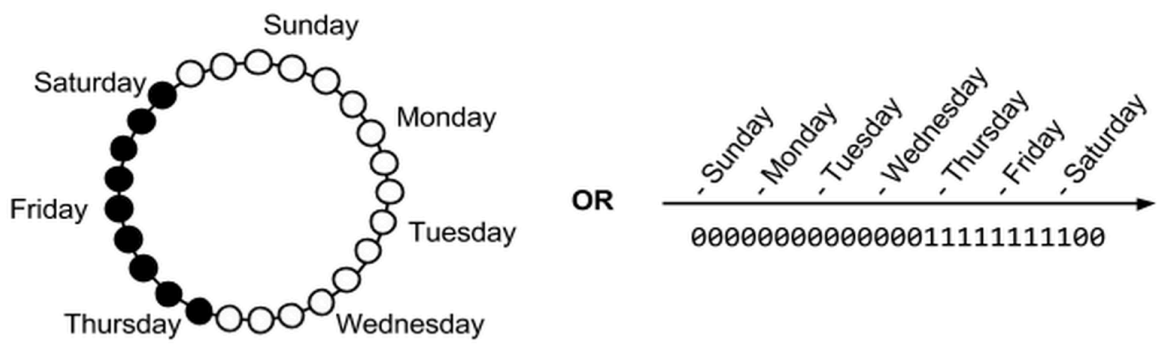

*Figure 4* The encoding for a date that falls on a Friday.

Both diagrams in Fig. 4 show a representation of the encoding for Friday. The left diagram shows how each day falls on a circle and overlaps with days before and after it. The right diagram shows the encoded representation of Friday along with annotations for where the center of each encoding falls. The encoding for Sunday would include bits both at the very beginning and at the very end (Fig. 5).

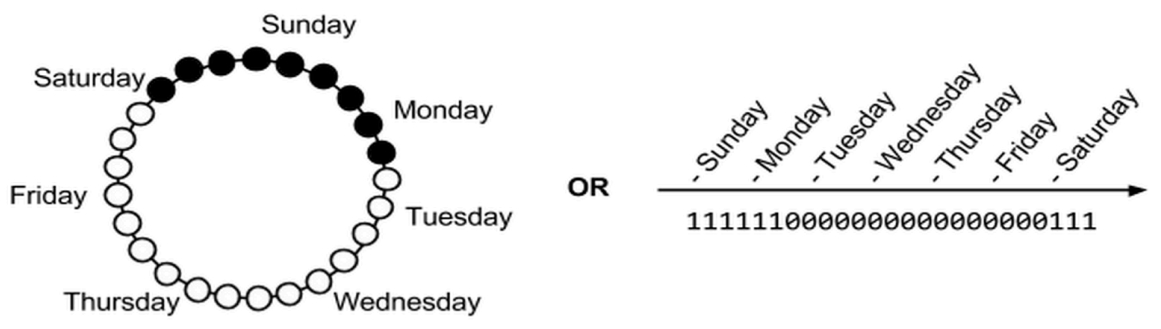

*Figure 5* The encoding for a date that falls on a Sunday.

Other numeric characteristics of dates and times that you may want to capture are:

- Month of the year
- Day of the month
- Time of day
- Minute of the hour

## 5. Example 3 – Encoding Geospatial Data

### 5.1 A Simple Encoder for Geospatial Data

This example shows how an encoder can capture geospatial data. The most obvious aspect is that locations close to each other should be considered similar while locations far apart should not be considered similar. To encode this meaning, we first have to determine the resolution that we want to encode. For this example, we will assume that we are using GPS coordinates that are accurate to about ten feet. And we will be doing two dimensional locations, although extending the encoding to three dimensions would be straight-forward.

The first step is to convert the GPS locations into a flat space that we can block off into ten-by-ten foot squares. Then we decide on an indexing system so that we can identify any section by integer X and Y coordinates[3].

Now we need to encode a location as a set of n bits with w active. For this example we will use n=100 and w=25 but real applications would want to use larger numbers, like n=1000. We first select the coordinates for the location we want to encode, say x=5 and y=10, and then identify the coordinates for the surrounding locations. These will form a square set of locations identified by having the coordinates 3<=x<=7 and 8<=y<=12. This process gives us 25 sets of x, y coordinates shown in gray here:

| ... | 8 | 9 | 10 | 11 | 12 | 13 | 14 | 15 | ... |
|---|---|---|---|---|---|---|---|---|---|
| 2 | | | | | | | | | |
| 3 | | | | | | | | | |
| 4 | | | | | | | | | |
| 5 | | | | | | | | | |
| 6 | | | | | | | | | |
| 7 | | | | | | | | | |
| 8 | | | | | | | | | |
| 9 | | | | | | | | | |
| ... | | | | | | | | | |

[3] The spherical mercator is one projection that can be used to transform GPS coordinates into two dimensional coordinates: https://en.wikipedia.org/wiki/Mercator_projection

Now we can use a deterministic hash function to map each pair of coordinates to the 100 bits in the encoding:

Hash(x, y) = ix,y

As explained earlier, by using a hash function we can use a fixed number of bits to represent any location in an unbounded space.

Because we are using a deterministic hashing function, we can compute these values when needed and do not need to store them. The encoding for one particular GPS coordinate ends up looking like this:

0000001001000000000101000000000000000000110001000000000000…

Note that the final encoding may have slightly fewer than 25 one bits due to hash collisions. When using sufficiently large numbers for n and w it is unlikely that this will happen and in any case it will not create problems in the HTM.

If we encode the position at x=6, y=10, most (20 out of 25) of the selected coordinates will overlap with the encoding for x=5, y=10, yielding the result we want of semantic overlap in SDRs.

### 5.2 A More Flexible Encoder for Geospatial Data

The previously described encoding method works well if you want as fine an encoding as the geospatial system allows. If this is 10 feet, then moving just 20 feet away will result in a slightly different SDR and moving a thousand feet will result in a completely different SDR. But this might not be want you want. You may want an encoder that encodes small differences when an object is moving slowly and larger distances when moving fast. For example if someone is walking, then we may want the encoder to represent changes in position as small as ten feet, but if someone is in a car moving at highway speeds, then ten feet differences in the direction of travel is not important. At high speed maybe positions within 200 feet are best represented by the same SDR and positions separated by 1,000 feet should be overlapping. This type of variable encoding is desired in many geospatial applications. Usually there is a way to design an encoder for any need. In this case the requirements can be met by forming an SDR using a subsample of all possible location bits and using the speed of the object to determine the coarseness of the sub-sampling.

The previously described geospatial encoding requires that the distance of two positions that you want to have overlap determine the number of active bits. You may not always want that many bits active. We can subsample from the range but need to determine which bits to subsample in a way that preserves the desired encoder properties. Specifically, we want data points that are close in space to have encodings that share bits.

If we randomly subsample 50% of the bits inside the radius, it is possible that the encoding for a nearby position shares no bits even though their radii overlap.

One way to solve this problem is to give a strict ordering to all coordinates that determines which bits to subsample. To do this, we can map each coordinate to a floating point number between 0.0 and 1.0 using a deterministic hashing function:

$$\text{Hash}(x, y) = w_{x,y}$$

Because we are using a deterministic hashing function, these weights can be computed when needed and do not need to be stored in memory. Here is what our previous encoding might look like if we wanted fifteen active bits in the encoding:

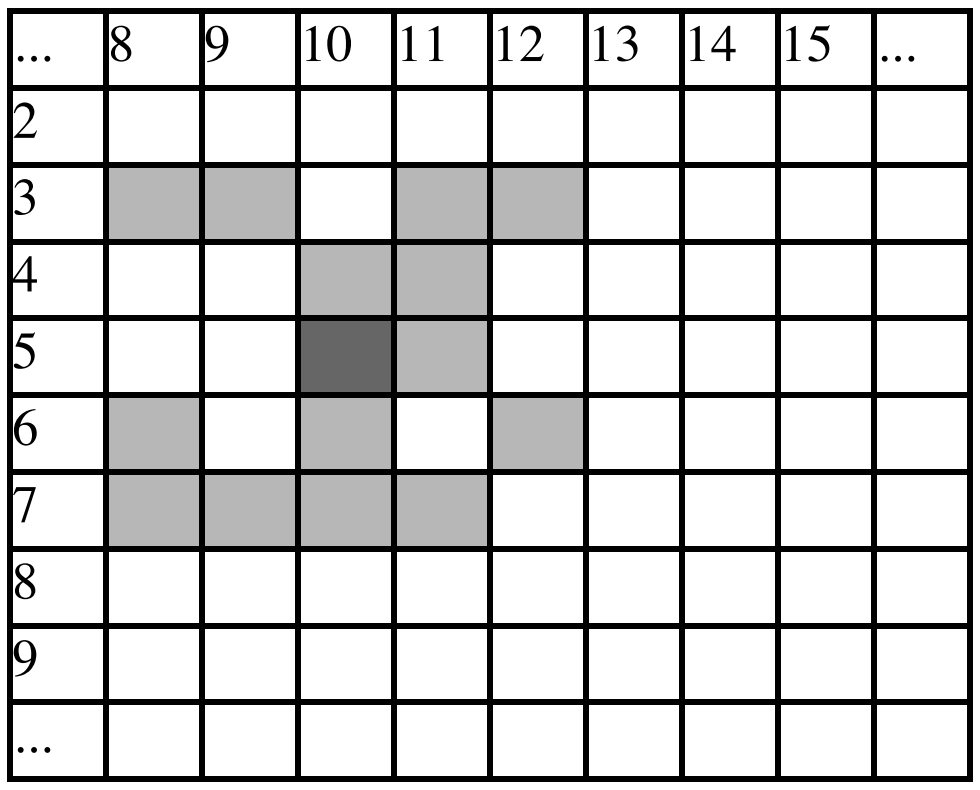

The weighting scheme helps preference the same coordinates so that the encodings for nearby positions select some of the same coordinates but it doesn't guarantee this. As a result, picking a sufficiently high number of total and active bits is necessary to ensure proper overlap.

## 5.3 Incorporating Speed in Geospatial Encoders

One promising application of coordinate or geospatial encoders is anomaly detection on people or vehicle positions. The speed of a person or vehicle is important to know when choosing how far apart positions should be before they no longer share overlap. One way to automate this process is to dynamically adjust the radius to select bits based on the current speed, changing the distance semantics to be relative to your speed rather than being an absolute metric.

The following figures (Fig. 6A, 6B and 6C) show representations of the encodings for three positions. The first two encodings are for positions for the entity while it is moving slowly, while the third represents a third position after the entity has sped up. Note that while the third position is much further away, it still shares a similar number of bits with the previous encoding.

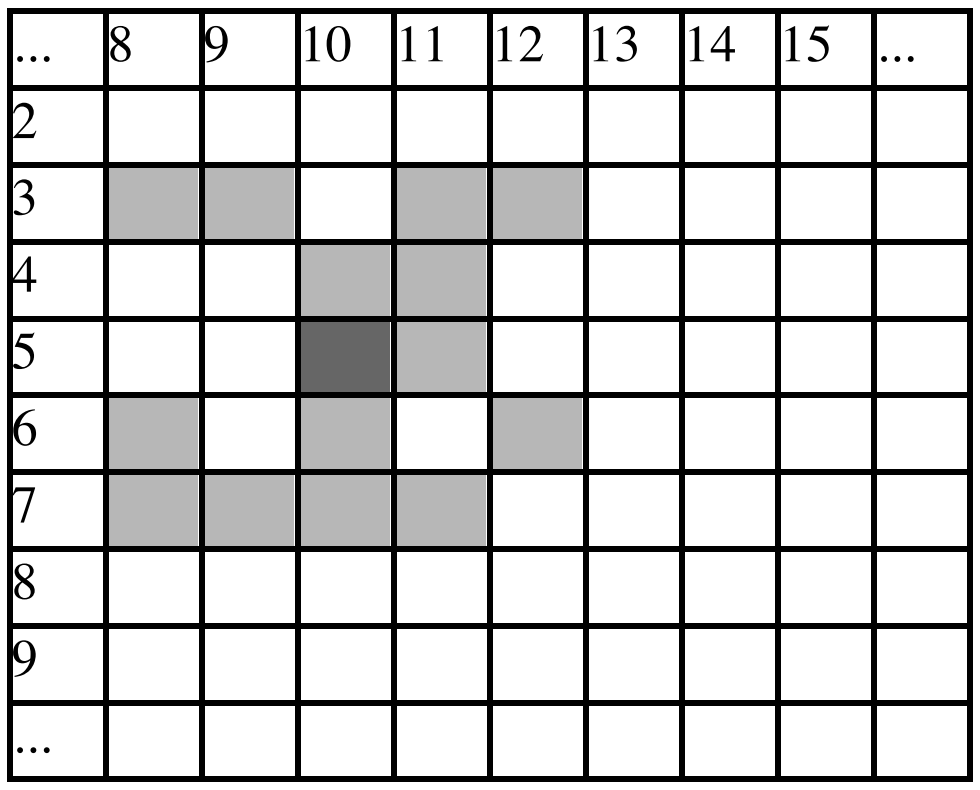

*Figure 6A* This is an encoding for a position in which the entity is moving slowly. The dark gray square represents the current location, X. The light gray squares are the ones chosen to be included in the representation from a radius of 2. These 15 squares have the highest weights of the 25 within the radius.

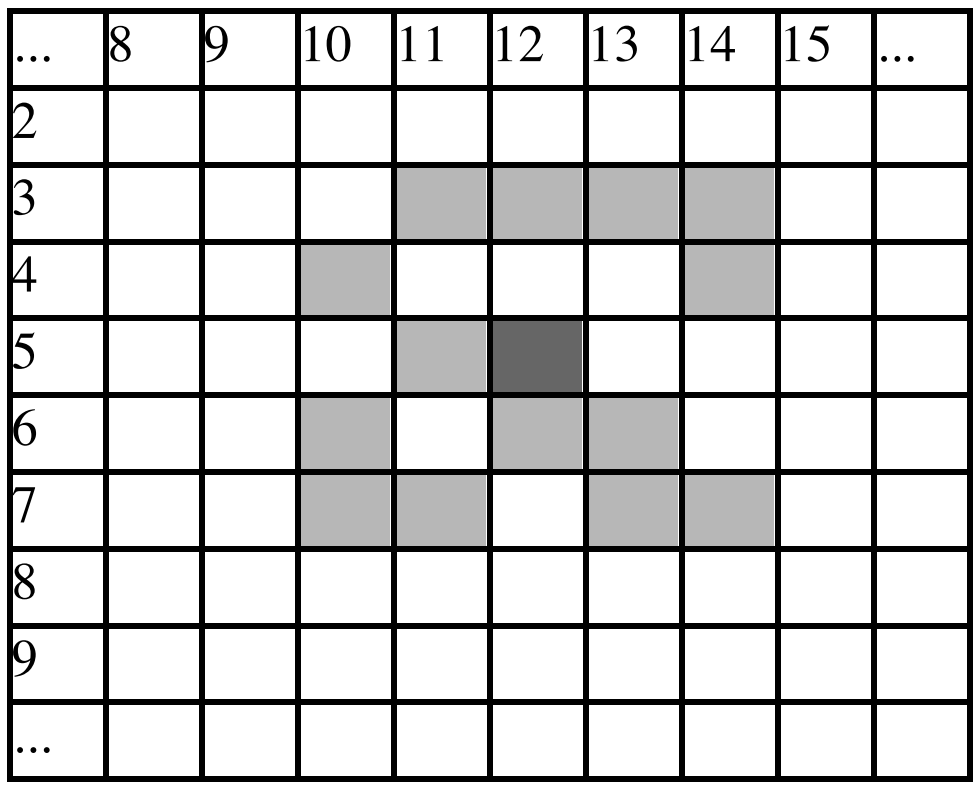

*Figure 6B* The encoding for a position nearby to X while the entity is still moving at the same speed. The entity has moved two coordinates to the right. For the coordinates that are within the radius for both positions, most of the same squares are selected because the weight for each square is fixed across encodings. And the radius is the same as X because the entity is moving at the same speed.

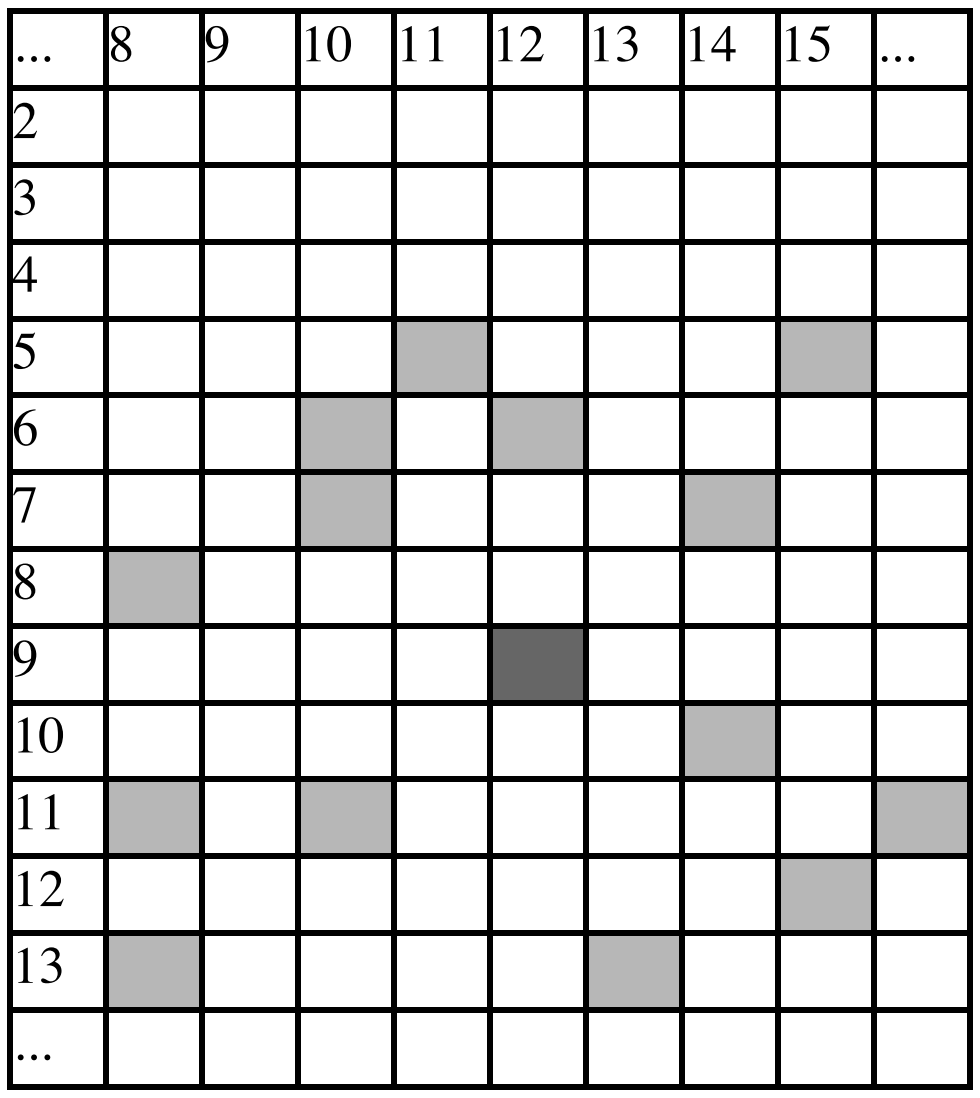

*Figure 6C* Here the entity has moved four coordinates down. Because the entity is moving faster, the radius has increased. The proportion of coordinates within the radius selected for the encoding is smaller because we are selecting the same number of bits, 15, out of a larger pool of 81 coordinates. Despite the new position being reasonably far away from Y and selecting a sparse set of the coordinates to include in the encoding, the weighting scheme ensures that there are still a handful of coordinates selected for the encodings of both Y and Z.

## 6. Example 4 – Encoding Natural Language

Words, sentences, and documents can be encoded into SDRs as well. There are many existing techniques for creating vector encodings of language, including one created by Cortical.io that produces SDRs. Their technique is described in detail in their recent white paper (Webber, 2015).

## 7. Choosing the Size of a New Encoder

Every encoder, no matter what it is representing, should create SDRs that have a fixed number of bits n and a fixed number of one bits w. How do you know what are good values for n and w? To maintain the properties that come from sparsity, w can't be a large fraction of n. But if w gets too small then we lose the properties that come from having a distributed representation. As a general rule, for numerical values w should be at least 20 to properly handle noise and subsampling and n should be at least 100 to provide enough resolution to distinguish between many numbers.

Here are some values that have been used successfully in practical applications.

- In one experiment looking at numeric metric values from servers, an optimization algorithm chose n = 134 and w = 21 as good values.
- In another experiment n = 2048 and w = 41 were found to be good values for a geospatial encoder.
- When encoding categories, w can be a higher proportion of n. For example, at one extreme if you encoded a binary value n could be 100 and w could be 50.

When creating a new encoder it is usually a good strategy to initially pick n and w using the broad guidelines. After the encoder is debugged and there are accuracy tests in place, then performance of the encoder can be tweaked by more carefully selecting n and w. A lot of the background for picking good values for n and w came from the work done by Ahmad and Hawkins on the properties of SDRs (Ahmad and Hawkins, 2016).

## 8. Encoding Multiple Values

Some applications require multiple values to be encoded for a single HTM model. The separate values can be encoded on their own and then concatenated to form the combined encoding. When doing this, it is important to keep the number of one bits, w, relatively similar for each of the individual value encoders so that one of the values does not dominate the representation. It is fine to have very different values for n, the number of total bits, for the individual encoders.

## 9. Conclusion

There are a number of encoders available that should cover the needs for most applications. And if you need to build an encoder for a new data type, there are a few simple rules that you can use to create the encoder:

1. Semantically similar data should result in SDRs with overlapping active bits.
2. The same input should always produce the same SDR as output.
3. The output should have the same dimensionality (total number of bits) for all inputs.
4. The output should have similar sparsity for all inputs and have enough one-bits to handle noise and subsampling.